%% file: main.tex
\documentclass{article}
\usepackage{spconf,amsmath,graphicx}

\usepackage[hyphens]{url}  
\usepackage{amsthm}
\usepackage{amssymb}

\usepackage[utf8]{inputenc} 
\usepackage[T1]{fontenc}    
\usepackage{hyperref}       
\usepackage{booktabs}       
\usepackage{amsfonts}       
\usepackage{nicefrac}       
\usepackage{microtype}      
\usepackage{times}
\usepackage{latexsym}
\usepackage{url}
\usepackage{mathtools}
\usepackage{CJKutf8}
\usepackage{array,multirow}
\usepackage[ruled]{algorithm2e}
\usepackage{comment}
\usepackage{color,soul}

\usepackage{subfigure}
\usepackage{appendix}
\usepackage{hhline}

\title{Improve Unsupervised Domain Adaptation with Mixup Training}
\name{Shen Yan$^{\dagger}$$^{\star}$\thanks{$^{\star}$Work done as an intern at Bosch Research North America.}, Huan Song$^{\ddagger}$, Nanxiang Li$^{\ddagger}$, Lincan Zou$^{\ddagger}$, Liu Ren$^{\ddagger}$}
\address{$^{\dagger}$ Michigan State University, East Lansing, MI \\
$^{\ddagger}$ Bosch Research North America, Sunnyvale, CA}
\begin{document}
%
\maketitle
\begin{abstract}
Unsupervised domain adaptation studies the problem of utilizing a relevant source domain with abundant labels to build predictive modeling for an unannotated target domain. Recent work observe that the popular adversarial approach of learning domain-invariant features is insufficient to achieve desirable target domain performance and thus introduce additional training constraints, e.g.\ cluster assumption. However, these approaches impose the constraints on source and target domains individually, ignoring the important interplay between them. In this work, we propose to enforce training constraints across domains using mixup formulation to directly address the generalization performance for target data. In order to tackle potentially huge domain discrepancy, we further propose a feature-level consistency regularizer to facilitate the inter-domain constraint. When adding intra-domain mixup and domain adversarial learning, our general framework significantly improves state-of-the-art performance on several important tasks from both image classification and human activity recognition.
\end{abstract}
\section{Introduction}
\input{intro.tex}
\section{Proposed Approach}
\input{proposed.tex}

\section{Experimental Results}
\input{results.tex}

\section{Conclusion}
\input{conclusion.tex}

\vfill\pagebreak

\bibliographystyle{IEEEbib}
\bibliography{refs}

\end{document}

%% file: intro.tex
Despite the success of deep learning based approaches in visual understanding and time-series analysis, they typically rely on abundant data and extensive human labeling. During deployment in real-world scenarios, they often face critical challenges when domain shifts occur and the labels under novel distributions are scarce or unavailable. It is crucial to develop highly effective domain adaptation scheme to transfer existing model trained on large-scale labeled data ({\it source domain}) to the related domain ({\it target domain}). In this work, we address the challenging {\it unsupervised domain adaptation} (UDA) problem, where the target domain is completely unannotated. 

A popular approach in UDA is learning indistinguishable representations between source and target domains through adversarial training. For example, the seminal DANN framework \cite{ganin2016domain} demonstrates that training with a domain discriminator directly minimizes the $\mathcal{H}$-divergence between domains \cite{ben2010theory} and hence induces domain-invariant representations. While several other existing work \cite{tzeng2017adversarial,long2018conditional,hoffman2018cycada} differ on the network and training paradigms, the overarching assumption is: when the domain discrepancy is addressed at the representation level, the trained source classifier is able to automatically achieve good performance on the target domain. However, recent research suggest that a classifier that performs well on both domains may be non-existent \cite{shu2018dirt,chen2019joint} and under this circumstance, solely relying on source classifier can lead to significant misclassifications in the target domain. 
This challenge motivates state-of-the-art approaches to seek additional training constraints during the adversarial learning process. VADA \cite{shu2018dirt} is the first approach proposed to minimize the conditional entropy of target domain, based on the cluster assumption commonly utilized in semi-supervised learning. On each domain individually, VADA further adds virtual adversarial training (VAT). In JDDA \cite{chen2019joint}, the authors propose metric-learning style losses on the source domain. By learning more compact and separable source features, it indirectly encourages more discriminative target domain features through domain alignment. 

We observe that the aforementioned studies employ training constraints in the chosen domain(s) independently, not jointly. This leaves the important interplay between the two domains unexplored and may significantly limit the potential of the training constraints. In this work, through the lens of the simple yet effective \textit{mixup} training \cite{zhang2017mixup}, we demonstrate that introducing training constraints \textit{across domains} can significantly improve the model adaptation performance. Denoting a pair of sample-label tuples as $(x_i,y_i)$, $(x_j, y_j)$, mixup generates augmented tuples as \cite{zhang2017mixup}:
\begin{gather}
    x'=\lambda x_i+(1-\lambda)x_j \nonumber\\
    y'=\lambda y_i+(1-\lambda)y_j \nonumber
\end{gather}
where $\lambda\in[0,1]$. By using the constructed $(x',y')$ for training, mixup encourages linear behavior of the model where linear interpolation in the raw data leads to linear interpolation of predictions.

Inspired by the recent advances in semi-supervised learning \cite{berthelot2019mixmatch}, we achieve mixup across domains through inferred labels on the target data. In this way, as opposed to training the classifier only with source labels, we are able to provide additional supervision also with interpolated (virtual) labels between domains. As the mixup training and domain adversarial training both progress, the model infers virtual labels with increased accuracy. This procedure can be critical to directly improve the generalization of the classifier when applied to target domain. Furthermore, to effectively enforce the linearity constraint under very large domain discrepancy, we develop a feature-level consistency regularizer to better facilitate the mixup training. Besides the inter-domain constraints, mixup can also be applied within each domain. The \textbf{i}nter- and \textbf{i}ntra-domain \textbf{m}ixup \textbf{t}raining constitute the proposed \textbf{IIMT} framework for enforcing multifaceted constraints to improve target domain performance. Our extensive evaluation on both visual recognition and human activity recognition demonstrate that IIMT significantly outperforms state-of-the-art methods on several important tasks.

%% file: proposed.tex

\begin{figure}[t]
\centering
\includegraphics[width=\linewidth]{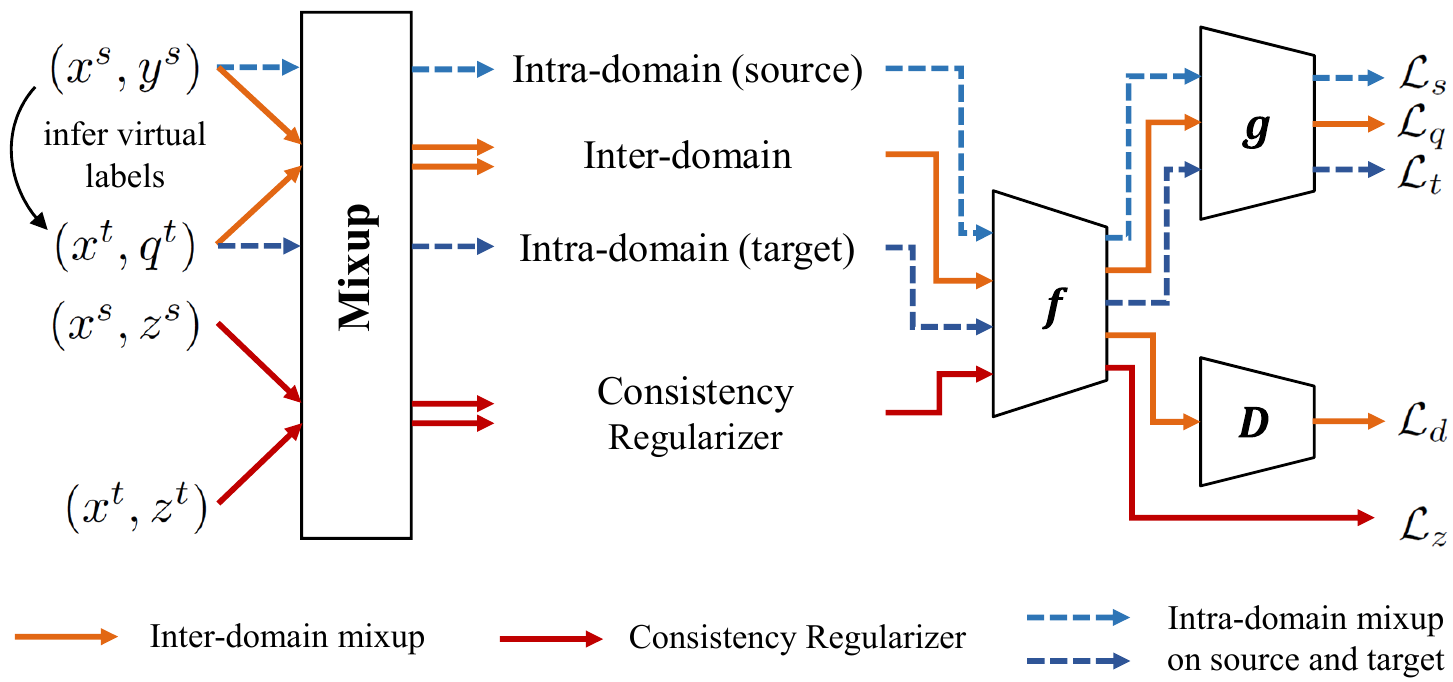}  
\caption{Overview of the proposed architecture highlighting mixup training. Note $z^s$ and $z^t$ come from the output of $f$.}
\label{fig:arch}
\end{figure}

The overview of IIMT framework is shown in Figure \ref{fig:arch}. We denote the labeled source domain as set $\{(x_i^s,y_i^s)\}_{i=1}^{m_s}\sim\mathcal{S}$ and unlabeled target domain as set $\{x_i^t\}_{i=1}^{m_t}\sim\mathcal{T}$. Here $y_i$ denotes one-hot labels. The overall classification model is denoted as $h_\theta:\mathcal{S}\mapsto\mathcal{C}$ with the parameterization by $\theta$. Following prominent approaches in UDA \cite{shu2018dirt,chen2019joint}, we consider the classification model as the composite of an embedding encoder $f_\theta$ and an embedding classifier $g_\theta$: $h=f\circ g$. Note that encoder is shared by the two domains. The core component in our framework is {\it mixup}, imposed both across domains ({\it Inter-domain} in Figure \ref{fig:arch}) and within each domain ({\it Intra-domain (source)} and {\it Intra-domain (target)} in Figure \ref{fig:arch}). All mixup training losses and the domain adversarial loss are trained end-to-end. 

\subsection{Inter-domain Mixup Training}
The key component in our proposed framework is the mixup training between source and target domains. This can be achieved after the target domain labels are inferred based on current classifier. In the training of $h$, mixup provides interpolated labels to enforce linear prediction behavior of the classifier across domains. Compared with training with source labels alone, this induces a simplistic inductive bias that can directly improve the generalization capability of the classifier for the target domain.

Mixup training requires sample labels in order to perform interpolation. We utilizes inferred labels as weak supervision for the target domain. Similar ideas have been shown to be highly effective in exploiting relevant unlabeled data in semi-supervised learning setting \cite{miyato2018virtual,berthelot2019mixmatch}. We follow \cite{berthelot2019mixmatch} by integrating data augmentation and entropy reduction into the virtual label generation process. First, we perform $K$ task-dependent stochastic augmentations (e.g.\ random crop and flipping for images and random scaling for time series) on each data sample to obtain transformed samples $\{\hat{x}_{i,k}\}_{k=1}^K$. Then, the target domain virtual labels are calculated as: $\bar{q}_i=\frac{1}{K}\sum_{k=1}^K h_\theta(\hat{x}_{i,k})$, and further normalized as: $q_i=\bar{q}_i^{\frac{1}{T}}/{\sum_c \bar{q}_{i,c}^{\frac{1}{T}}}$. Here $T$ denotes the softmax temperature and $c$ is the class index. That is, the class predictions over the $K$ augmented inputs are first averaged to constitute $\bar{q}_i$ and further sharpened to form the virtual labels $q_i$. Using smaller value $T<1$ produces sharper predicted distributions and helps to reduce the conditional entropy when $q_i$ is used in training.


Given a pair of source and target samples, $(x_i^s,x_i^t)$, taken from their corresponding batches, label-level mixup to enforce the linearity consistency across domains can be defined as:
\begin{gather}
x_i^{st}=\lambda' x_i^s+(1-\lambda')x_i^t \label{eq:sample-mixup}\\
q_i^{st}=\lambda' y_i^s+(1-\lambda')q_i^t \label{eq:label-mixp} \\
\mathcal{L}_q=\frac{1}{B}\sum_i H(q_i^{st}, h_\theta(x_i^{st})) \label{eq:consis-label}
\end{gather}
where $B$ denotes the batch size, $H$ denotes the cross-entropy loss and the mixup weighting parameter is selected according to: $\lambda\sim \text{Beta}(\alpha,\alpha)$ and $\lambda'=\max(\lambda,1-\lambda)$. Here $\text{Beta}$ refers to beta distribution with the shared shape parameter $\alpha$. When $\alpha$ is set to closer to $1$, there is a larger probability to choose medium value of $\lambda$ from range $[0,1]$, leading to higher level of interpolation between the two domains. Note that $\lambda'$ is always over $0.5$ to ensure the source domain is dominant. Similarly the mixup dominated by the target domain can be generated via switching $x^s$ and $x^t$ in Eq. (\ref{eq:sample-mixup}), which forms $(x^{ts},q^{ts})$. With $(x^{ts},q^{ts})$ we utilize the mean square error (MSE) loss as it is more tolerant to false virtual labels in the target domain. 


\subsubsection{Consistency Regularizer}
In the scenario of very large domain discrepancy, the linearity constraint imposed by inter-domain mixup could be less effective. Specifically, when the heterogeneous raw inputs are interpolated in Eq. (\ref{eq:sample-mixup}), forcing the model $h$ to produce correspondingly interpolated predictions becomes significantly harder. At the same time, the joint training with domain adversarial loss (details in next section) for feature-level domain confusion can add to the training difficulty.

These challenges motivate us to design a consistency regularizer for the latent features to better facilitate inter-domain mixup training. Denoting $\mathcal{Z}$ as the embedding space induced by $f$ and $z\in\mathcal{Z}$, we define the regularizer term as:
\begin{gather}
z_i^{st}=\lambda' f_\theta(x_i^s)+(1-\lambda')f_\theta(x_i^t) \label{eq:feat-mixup} \\
\mathcal{L}_z=\frac{1}{B}\sum_i\left\|z_i^{st}-f_\theta(x_i^{st})\right\|_2^2 \label{eq:consis-feat}
\end{gather}
where $x_i^{st}$ is from Eq. (\ref{eq:sample-mixup}). That is, we push the mixed feature closer to the feature of the mixed input through MSE loss between the two vectors. In this way, we impose the linearity constraint to be enforced also at the feature level. The efficacy of this regularizer is that when Eq. (\ref{eq:consis-feat}) is enforced and $z_i^{st}, f_\theta(x_i^{st})$ are passed through the shallow classifier $g$, the linearity in the model prediction becomes much easier to be satisfied. Note that similar to the handling of Eq. (\ref{eq:sample-mixup}), $x_i^s$ and $x_i^t$ can also be switched in Eq. (\ref{eq:feat-mixup}) to form $z_i^{ts}$. We omitted the loss term on that for clarity.

\subsubsection{Domain Adversarial Training}
Our last component in inter-domain mixup is employing standard domain adversarial training to reduce the domain discrepancy. Here we restrict our implementation to the more fundamental DANN framework \cite{ganin2016domain}, as an attempt to focus on evaluating the mixup linearity constraints. In DANN, a domain discriminator and the shared embedding encoder (generator) are trained under the adversarial objective such that the encoder learns to generate domain-invariant features. Denote the domain discriminator as $D:\mathcal{Z}\mapsto (0,1)$, where $0/1$ annotates binary domain label. We define the domain adversarial loss on the mixed source and target samples:
\begin{equation}
    \mathcal{L}_d=\frac{1}{B}\sum_i\ln D(f_\theta(x_i^{st}))+\ln(1-D(f_\theta(x_i^{st}))) \label{eq:adv}
\end{equation}

\subsection{Intra-domain Mixup Training}
Given the source labels and target virtual labels, mixup training can be performed within each domain too. Since samples in the same domain follow the similar distribution, there is no need to apply feature-level linearity. We therefore employ only label-level mixup training for both domains and define their corresponding losses:
\begin{gather}
{x_i^s}'=\lambda' x_i^s+(1-\lambda')x_j^s \nonumber \\
{y_i^s}'=\lambda' y_i^s+(1-\lambda')y_j^s \nonumber \\
\mathcal{L}_s=\frac{1}{B}\sum_i H({y_i^s}',h_\theta({x_i^s}')) \label{eq:src-loss}\\
{x_i^t}'=\lambda' x_i^t+(1-\lambda')x_j^t \nonumber \\
{q_i^t}'=\lambda' q_i^t+(1-\lambda')q_j^t \nonumber \\
\mathcal{L}_t=\frac{1}{B}\sum_i\left\|{q_i^t}'-h_\theta({x_i^t}')\right\|_2^2 \label{eq:tgt-loss}
\end{gather}
Although within-domain mixup is intuitive as a data augmentation strategy, it is particularly useful for UDA. As discussed in \cite{shu2018dirt}, the minimization of conditional entropy without locally-Lipschitz constraints can results in abrupt prediction changes in the vicinity of data samples. In \cite{shu2018dirt}, virtual adversarial training \cite{miyato2018virtual} is utilized to enforce prediction smoothness in neighborhood. Differently, we find that intra-domain mixup training is able to achieve the same objective. We provide more empirical details in the experiments to demonstrate.

\subsection{Overall Architecture}
We illustrate the overall architecture in Figure \ref{fig:arch}. Summarizing all previous components, we arrive at the final loss:
\begin{equation}
\mathcal{L}=w_q\mathcal{L}_q+w_d\mathcal{L}_d+w_z\mathcal{L}_z+w_s\mathcal{L}_s+w_t \mathcal{L}_t
\end{equation}
Since $\mathcal{L}_t$ only involves virtual labels, it could be easily affected by the uncertainty in target domain. We set a linear schedule for $w_t$ in training, from $0$ to a predefined maximum value. From initial experiments, we observe that the algorithm is robust to other weighting parameters. Therefore we only search $w_t$ while simply fixing all other weightings to $1$. 

%% file: results.tex
Our evaluation consists of image classification and human activity recognition (HAR). For the first part, we consider visual domain adaptation experiments commonly used in the UDA literature. They are constructed on MNIST, MNIST-M, Street View House Numbers (SVHN), Synthetic Digits (SYN DIGITS), CIFAR-10 and STL-10. For HAR experiments, we evaluate on OPPORTUNITY \cite{ordonez2016deep} and WiFi \cite{yousefi2017survey} datasets. We use A $\rightarrow$ B to denote the domain adaptation task with source domain A and target domain B. 

The architecture of the embedding encoder $f$ is task-specific: we use similar architectures to \cite{shu2018dirt} for visual recognition and WiFi datasets; we employ state-of-the-art DeepConvLSTM architecture \cite{ordonez2016deep} for OPPORTUNITY. For the domain discriminator $D$, we forward the extracted feature to a simple fully-connected layer ($x-128-\text{ReLU}-1$). We adopt shallow fully-connected networks for embedding classifier $g$.

\begin{table}[t]
\centering
\resizebox{\linewidth}{!}{
\begin{tabular}{cccccc}
\hline

Source & \textbf{MNIST} & \textbf{SVHN} & \textbf{SYN} & \textbf{CIFAR} & \textbf{STL} \\
Target & \textbf{MNISTM} & \textbf{MNIST} & \textbf{SVHN} & \textbf{STL} & \textbf{CIFAR} \\ \hline

DAN \cite{long2015learning} & 76.9 & 71.1 & 88.0 & - & - \\
DANN \cite{ganin2016domain} & 81.5 & 71.1 & 90.3 & - & - \\
DRCN \cite{Ghifary2016DeepRN} & - & 82.0 & - & 66.4 & 58.7 \\
ATT \cite{saito2017asymmetric}  & 94.2 & 86.2 & 92.9 & - & - \\
$\Pi$-model \cite{French2017SelfensemblingFV} & - & 92.0 & 94.2 & 76.3 & 64.2 \\
JDDA \cite{chen2019joint} & 88.4 & 94.2 & - & - & - \\
\hline
Source-Only & 62.5 & 72.6 & 88.1 & 75.9 & 61.8 \\ 
VADA \cite{shu2018dirt} & 97.7 & 97.9 & 94.8 & 80.0 & 73.5 \\
DIRT-T \cite{shu2018dirt} & 98.9 & \textbf{99.4} & 96.1 & 80.0 & 75.3 \\
\textbf{IIMT} & \textbf{99.5} & 97.3 & \textbf{97.0} & \textbf{83.1} & \textbf{81.6} \\
\hline
\end{tabular}
}
\caption{Test set accuracy ($\%$) on visual UDA benchmarks.}
\label{tab.visual.domain.experiments}
\end{table}

\begin{table}[t]
\centering
\resizebox{\linewidth}{!}{
\begin{tabular}{cccccccc}
\hline
\multicolumn{1}{c}{\textbf{Method}} &
\textbf{\begin{tabular}[c]{@{}c@{}} 1 $\rightarrow$ 2 \\ \end{tabular}} & \textbf{\begin{tabular}[c]{@{}c@{}} 1 $\rightarrow$ 3 \\ \end{tabular}} & \textbf{\begin{tabular}[c]{@{}c@{}} 2 $\rightarrow$ 1 \end{tabular}} & \textbf{\begin{tabular}[c]{@{}c@{}} 2  $\rightarrow$ 3 \\ \end{tabular}} & \textbf{\begin{tabular}[c]{@{}c@{}} 3 $\rightarrow$ 1 \\ \end{tabular}} &
\textbf{\begin{tabular}[c]{@{}c@{}} 3 $\rightarrow$ 2 \\ \end{tabular}} &
\textbf{\begin{tabular}[c]{@{}c@{}} Avg \\ \end{tabular}}\\   \hline

Source-Only                 &	0.652 &	0.640 &	0.696 &	0.637 &	 0.659 &	0.631 &	0.652 \\ 
DANN &	0.768 &	0.731 &	0.785 &	0.694 &	 0.746 &	0.725 &	0.741 \\ 
VADA     &	0.776 &	0.747 & 0.797 &	0.734 &	0.726 &	0.720 &  0.750 \\
\textbf{IIMT}     & \textbf{0.809} &	\textbf{0.780}	& \textbf{0.813} &	\textbf{0.745} &	\textbf{0.831} &	\textbf{0.787} &	\textbf{0.794} \\
\hline
\end{tabular}
}
\caption{Test set weighted F1 score on OPPORTUNITY. }
\label{tab.oppo.experiments}
\end{table}

\begin{table}[t]
\centering
\scalebox{0.77}{
\begin{tabular}{cc}
\hline
\multicolumn{1}{c}{\textbf{Method}} &
\textbf{\begin{tabular}[c]{@{}c@{}} Room A $\rightarrow$ Room B \\ \end{tabular}} \\   \hline 
Source-Only & 36.4   \\ 
DANN & 38.7 \\
VADA \cite{shu2018dirt} & 53.0 \\
DIRT-T \cite{shu2018dirt} & 53.0 \\
\textbf{IIMT} & \textbf{60.6} \\
\hline
\end{tabular}
}
\caption{Test set accuracy ($\%$) on WiFi HAR UDA task.}
\label{tab.non-visual-domain.wifi.experiments}
\end{table}

\subsection{Evaluation on Visual Recognition}
Our benchmarking results on common visual domain adaptation tasks are summarize in Table \ref{tab.visual.domain.experiments}. For digits classification UDA tasks, our proposed approach outperforms the state-of-the-art results achieved by DIRT-T for MNIST $\rightarrow$ MNISTM and SYN DIGITS $\rightarrow$ SVHN. Note that DIRT-T accuracy is close to $100\%$ and further improvement is very valuable. For SVHN $\rightarrow$ MNIST, our performance is on par with VADA while only scoring lower than DIRT-T. 

For object recognition UDA tasks CIFAR $\leftrightarrow$STL, the two adaptation directions have different degrees of difficulty. Since STL is much smaller, a model trained on it without any adaptation performs badly on the target domain. For both directions, our algorithm significantly outperforms state-of-the-art methods: in CIFAR $\rightarrow$ STL, we achieve $3.1\%$ margin-of-improvement; for STL $\rightarrow$ CIFAR, we achieve $8.1\%$ and $6.3\%$ margin-of-improvement on VADA and DIRT-T respectively. Note that DIRT-T had a remarkable improvement of $11\%$ over $\Pi$-model for STL $\rightarrow$ CIFAR. The capability to achieve significant further improvement demonstrates that, the proposed approach is highly effective in exploiting the target domain unlabeled data, even when source domain labels are limited.

\subsection{Evaluation on Human Activity Recognition}
In building HAR system for fitness monitoring and assisted living \cite{mohr2017personal,song2018optimizing}, acquiring training data requires careful system setup, long-term human subjects involvement and laborious labeling efforts. Since repeating such procedure for new sensing environment is prohibitive in practice, UDA becomes crucial in the deployment of practical HAR systems. We conduct experiments on this problem with both IMU sensor and WiFi based HAR datasets. Note that \textit{domain} refers to human subject in the first application and corresponds to sensing room in the second problem. 

Sensor-based HAR is performed on the public OPPORTUNITY repository containing sensor signals of sporadic gestures. Based on the first $3$ subjects, we construct $6$ cross-subject UDA experiments to explicitly investigate the influence of user (domain) change to the classifier accuracy. The sensor signals are segmented using a $0.8$s moving window with half overlapping. Due to the imbalanced class distribution (caused by the \textit{null} class), we use the weighted F1 score as the evaluation metric. The experimental results are summarized in Table \ref{tab.oppo.experiments}. First, we observe that without domain adaptation, all classifiers have inferior performance on the target subject, thus evidencing the necessity of UDA in HAR. Second, possibly due to the high class imbalance, VADA performance is only slightly better than conventional adversarial training. On the other hand, our proposed method significantly outperforms both approaches, achieving an averaged improvement over $0.04$.

\begin{table}[t]
\centering
\scalebox{0.77}{
\begin{tabular}{ccc}
\hline
\textbf{Method} & \textbf{STL $\rightarrow$ CIFAR} & \textbf{1 $\rightarrow$ 2} \\   \hline 
Source-Only & 61.8 & 0.652  \\ 
DANN  & 63.4 & 0.768 \\
Add intra-domain $\mathcal{L}_s$, $\mathcal{L}_t$  &  75.7 &  0.787\\
Add inter-domain $\mathcal{L}_q$ & 79.2 & 0.798 \\
Add inter-domain $\mathcal{L}_z$ & 81.6 & 0.809 \\
\hline
\end{tabular}
}
\caption{Ablation experiments for STL $\rightarrow$ CIFAR and OPPORTUNITY 1 $\rightarrow$ 2 tasks. Each row corresponds to adding the specified component(s) to the previous row.}
\label{tab.visual-domain.ablation.experiments}
\end{table}
%
%
%
%
%

We conduct experiments on the same WiFi activity recognition dataset \cite{yousefi2017survey} used in \cite{shu2018dirt}. In here the CSI from commercial WiFi systems was collected in $2$ different rooms. We conduct the adaptation experiment from Room A to B in recognizing the $7$ physical activities such as walk, fall, pickup etc. The $90$-dimensional CSI time series are temporally segmented with a $1$s moving window and $0.2$s stride. As can be seen from the experimental results in Table \ref{tab.non-visual-domain.wifi.experiments}, our approach significantly outperforms VADA/DIRT-T.

Finally, we perform an ablation analysis in order to study the contribution of each mixup component and the results are listed in Table \ref{tab.visual-domain.ablation.experiments}. We observe that each proposed component is effective in achieving performance gain. Importantly, the performance with intra-domain mixup training is already slightly better than VADA (see Table \ref{tab.visual.domain.experiments} and \ref{tab.oppo.experiments}). This demonstrates that intra-domain mixup is as effective as VAT \cite{shu2018dirt} in enforcing the locally-Lipschitz constraint. While \cite{shu2018dirt} utilizes VAT on each domain, we additionally perform cross-domain mixup. The two inter-domain components in Table \ref{tab.visual-domain.ablation.experiments} gain collective improvement of $6\%$ for STL $\rightarrow$ CIFAR and $0.02$ for 1 $\rightarrow$ 2.
%
%

%% file: conclusion.tex
This paper studies imposing cross-domain training constraints for domain adaptation through the lens of mixup linearity. A consistency regularizer is proposed to facilitate inter-domain mixup training in the presence of large domain discrepancy. The general IIMT framework incorporating both inter- and intra-domain mixup training outperforms state-of-the-art methods in diverse application areas. 